\documentclass[10pt,twocolumn,letterpaper]{article}

\usepackage[pagenumbers]{cvpr} 

\definecolor{cvprblue}{rgb}{0.21,0.49,0.74}
\usepackage[pagebackref,breaklinks,colorlinks,allcolors=cvprblue]{hyperref}

\newcommand\blfootnote[1]{%	
  \begingroup
  \renewcommand\thefootnote{}\footnote{#1}%
  \addtocounter{footnote}{-1}%
  \endgroup
}

\usepackage[capitalize]{cleveref}

\usepackage{multirow}
\usepackage{makecell} 
\usepackage{animate}
\usepackage{graphicx}
\usepackage{xcolor}
\usepackage[misc]{ifsym}
\usepackage{footnote}
\crefname{section}{Sec.}{Secs.}
\Crefname{section}{Section}{Sections}
\Crefname{table}{Table}{Tables}
\crefname{table}{Tab.}{Tabs.}
\newcommand{\name}{{Imagine360}}

\title{\name: Immersive 360 Video Generation from Perspective Anchor}

\author{
Jing Tan$^1$$^*$, Shuai Yang$^{2,4}$$^*$, Tong Wu$^1$\textsuperscript{\Letter}, Jingwen He$^1$, Yuwei Guo$^1$, \\ 
Ziwei Liu$^3$, Dahua Lin$^{1,4}$\textsuperscript{\Letter}\\
        $^1$ The Chinese University of Hong Kong \quad
        $^2$ Shanghai Jiao Tong University  \\
        $^3$ Nanyang Technological University \quad
        $^4$ Shanghai AI Laboratory \\
        \tt\small
        \href{https://ys-imtech.github.io/projects/Imagine360}{https://ys-imtech.github.io/projects/Imagine360}
}

\begin{document}

%%% On page 1
\twocolumn[{
\renewcommand\twocolumn[1][]{#1}
\maketitle
\begin{center}
    \centering
    \vspace{-20pt}
    \animategraphics[loop,controls,width=1.0\linewidth]{8}{figures/teaser/frame_}{1}{32}
  \captionof{figure}{\small
  \textbf{Overview of~\name.} ~\name~lifts standard perspective video into $360^\circ$ video, enabling dynamic scene experience from full 360 degrees. Compared to Follow-Your-Canvas, {which focuses only on \textbf{perspective} visual and motion patterns,} our approach achieves more plausible \textbf{spherical} video patterns. \textbf{\textit{\color{purple}Best viewed with Acrobat Reader for the \underline{animated} 360 videos.}}
  }
  \label{fig:teaser}
  % \vspace{-8pt}
\end{center}
}]

%%% One page 2
% \begin{figure*}[h]
%   \centering
%   \animategraphics[loop,controls,width=1.0\linewidth]{8}{figures/teaser/frame_}{1}{32}
%   \caption{\textbf{Overview of~\name.} ~\name~lifts standard perspective video into $360^\circ$ video, enabling dynamic scene exploration from full 360 degrees. Our approach achieves more plausible spherical video patterns compared to Follow-Your-Canvas. \textbf{\textit{\color{purple}Better viewed with Acrobat Reader for the animated 360 videos.}}}
%   \label{fig:teaser}
%   \vspace{-8pt}
% \end{figure*}

% \includemedia[
%   activate=onclick,
%   width=300pt, height=200pt,
%   addresource=figures/teaser.mp4,
%   flashvars={
%      source=figures/teaser.mp4
%     &autoPlay=true
%   }
% ]{}{VPlayer.swf}

% \includemedia[
%   width=0.4\linewidth,
%   height=0.3\linewidth,
%   activate=pageopen,
%   addresource=figures/teaser.mp4,
%   flashvars={source=figures/teaser.mp4}]{}{VPlayer.swf}

% \begin{figure}[h]
%   \centering
%   \movie[width=0.8\linewidth, height=0.6\linewidth, showcontrols]{}{figures/teaser.mp4}
%   \caption{teaser}
% \end{figure}

% \includemedia[
%     width=0.6\linewidth,
%     height=0.7\linewidth,
%     keepaspectratio,
%     activate=pageopen,
%     playbutton=fancy,
%     addresource=figures/teaser.mp4,
%     flashvars={source=figures/teaser.mp4&autoPlay=true}
%     ]{}{VPlayer.swf}
\blfootnote{\small *~: Equal contribution. \Letter~: Corresponding author.}

\begin{abstract}
$360^\circ$ videos offer a hyper-immersive experience that allows the viewers to explore a dynamic scene from full 360 degrees. 
To achieve more user-friendly and personalized content creation in $360^\circ$ video format, we seek to lift standard perspective videos into $360^\circ$ equirectangular videos. To this end, we introduce~\textbf{\name}, the first perspective-to-$360^\circ$ video generation framework that creates high-quality $360^\circ$ videos with rich and diverse motion patterns from video anchors.
\name~learns fine-grained spherical visual and motion patterns from limited $360^\circ$ video data with several key designs. 
\textbf{1)} Firstly we adopt the dual-branch design, including a perspective and a panorama video denoising branch to provide local and global constraints for $360^\circ$ video generation, with motion module and spatial LoRA layers fine-tuned on extended web $360^\circ$ videos.
\textbf{2)} Additionally, an antipodal mask is devised to capture long-range motion dependencies, enhancing the reversed camera motion between antipodal pixels across hemispheres.
\textbf{3)} To handle diverse perspective video inputs, we propose elevation-aware designs that adapt to varying video masking due to changing elevations across frames.
Extensive experiments show~\name~achieves superior graphics quality and motion coherence among state-of-the-art $360^\circ$ video generation methods. We believe~\name~holds promise for advancing personalized, immersive $360^\circ$ video creation. 
\end{abstract}    
\section{Introduction}
\label{sec:intro}
Imagine taking on a journey in the heart of a bustling city, a serene beach, or a cherished place of your own. As you turn your head, you will see the evolving world at new viewpoints, allowing for dynamic experience in full 360 degrees. $360^\circ$ video offers an interactive, immersive viewing experience that creates a living, breathing world as if the viewer is part of the experience. 
With the rapid development of head-mounted spatial computing systems, the demand for $360^\circ$ videos is increasing, driven by extensive applications across entertainment, education, and communication. 

Recent advancements in $360^\circ$ video generation have focused on text-guided~\cite{360dvd} and image-guided~\cite{4k4dgen} models.
While these methods produce plausible $360^\circ$ videos, they require panoramic optical flow~\cite{360dvd} or high-quality panoramic images~\cite{4k4dgen} as guidance, which are hard to obtain for average users. 
In contrast, perspective videos are much more accessible, easily captured with smartphone cameras, or generated by advanced video generation models. To enable more user-friendly and personalized $360^\circ$ video creation, we propose a new task: perspective-to-$360^\circ$ video generation, which transforms perspective video inputs into $360^\circ$ equirectangular videos.
Specifically, we take a perspective video with narrow FOV as the anchor video, project it to a $360^\circ \times 180^\circ$ FOV video canvas with perspective-to-equirectangular (P2E) mapping and create the unknown surrounding pixels. With dynamic and diverse motion condition information from anchor videos, this video-based guidance offers a more effective paradigm to generate high-quality 360° videos with richer and more complex motions compared to text- or image-based guidance.

One relevant task is video outpainting, which aims to fill in missing regions outside the edges of video frames in a larger canvas, typically in the perspective domain with fixed video masking across frames. 
Simply applying standard video outpainting methods does not achieve satisfactory results, as our perspective-to-$360^\circ$ video generation exhibits more challenges.
First, due to the large domain gap between perspective and panorama videos, learning the spherical visual and motion patterns requires sophisticated design when trained on limited panorama video data.
Second, as videos exhibit different elevation angles across frames, after the P2E mapping, the masking continuously changes in shape, size, and location, shown in \cref{fig:teaser}, requiring elevation-aware designs for robust generation.

To address these challenges, we introduce~\name, the first perspective-to-$360^\circ$ video generation framework, that creates high-quality panoramic videos from perspective anchor videos. 
Directly fine-tuning a pre-trained video diffusion model from the perspective domain proves inadequate for generating high-quality panoramic patterns. 
To tackle this issue, we adapt the \textbf{dual-branch design}~\cite{panfusion} from image-level to conditioned video generation. 
Specifically, we devise two parallel branches for perspective and panoramic video denoising, with each branch comprising spatial layers, motion modules, and cross-domain attention layers. 
The two branches are tightly coupled via bidirectional direct mapping (P2E and E2P) through cross-domain attention, ensuring the generation of plausible spherical motion patterns.
However, direct mapping alone is insufficient for capturing long-range dependencies essential for $360^\circ$ videos. 
It often fails to account for a critical characteristic of panoramic videos, where each pixel undergoes reverse camera motion of its antipodal counterpart. 
Hence, we improve the cross-domain attention with \textbf{antipodal mask}, to facilitate information exchange between each pixel and its \textit{antipodal} counterparts on the opposite hemisphere. 
In this regard, the receptive field of each pixel is extended from the directly mapped neighborhood to the antipodal neighborhood, making it easier to learn long-range panoramic motion dependencies. 

Moreover, taking general videos as anchors for personalized creation requires the framework to cater to different yaw and pitch angles in the anchor video. 
Benefiting from the $360^\circ$ close-loop property, we can assume the viewer rotates with the camera, and without loss of generality, only consider the influence of the changing pitch angles. 
In practice, we incorporate \textbf{elevation-aware designs}, including elevation-aware data sampling in training and an elevation estimation module in inference, to ensure robust generation for customized video inputs.
With the three key designs,~\name~pioneers the end-to-end, high-quality $360^\circ$ video generation from perspective video inputs.

Extensive experiments show that our model generates $360^\circ$ videos with the best frame quality and motion quality among state-of-the-art $360^\circ$ video generation methods. We also show a bonus advantage of our pipeline for achieving superior panorama image outpainting results. We believe that~\name~is able to lead the $360^\circ$ video generation community in creating personalized, hyper-immersive experiences for downstream applications.

\section{Related Works}
\label{sec:related}

\subsection{Diffusion Models}
Diffusion models \cite{ddpm, songyang1, songyang2} have achieved remarkable success in image generation \cite{sd, sdxl, sd3}, leading to advancements in video diffusion models \cite{cogvideox, animatediff, videoldm, modelscope, videocrafter1, videocrafter2, imagenvideo}. 
The first video diffusion model (VDM) \cite{vdm} adopts a space-time factorized U-Net in pixel space to model low-resolution videos. Imagen-Video \cite{imagenvideo} proposes to use cascaded DMs for generating high-definition videos. Subsequent research \cite{videoldm, animatediff, videocrafter1, videocrafter2, modelscope} adapts existing text-to-image (T2I) models to text-to-video (T2V) models by incorporating temporal layers, including both convolution and attention layers. More recently, several works \cite{t2x, cogvideox} directly use 3D full attention to model space-time information for more unified video representation. On the other hand, image-to-video (I2V) \cite{i2vgenxl, dynamicrafter, sparsectrl, svd, pia, t2i_adaptor} has arisen great attention as it enables more precise control on video generation. Some works achieve I2V by incorporating the image condition into the pretrained T2V models and finetuning newly added modules \cite{dynamicrafter, sparsectrl} or the inherited weights \cite{svd, i2vgenxl}, while plug-to-play methods \cite{pia, t2i_adaptor} aims to turn any text-to-image models into image animators.

\subsection{Video Outpainting}
Video Outpainting aims to fill in the missing regions at the edges of source videos. Compared to advanced image-level outpainting, video-level outpainting remains under-explored due to its challenges in maintaining both spatial and temporal fidelity and consistency. Recent video outpainting methods leverage diffusion to generate high-quality pixels in the missing regions. M3DDM~\cite{m3ddm} proposes a frame-guided Masked 3D diffusion model and a coarse-to-fine inference strategy to tackle artifact accumulation in long video outpainting. MOTIA~\cite{motia} employs a per-case optimization strategy to learn the data-specific patterns of source video for better outpainting quality. Follow-Your-Canvas~\cite{followyourcanvas} divides the canvas into multiple windows and achieves outpainting of different sizes and resolutions by merging each outpainted window. These methods focus on handling perspective video outpainting with fixed masking in each frame.
Despite their appealing outpainting results, it remains difficult for them to handle perspective-$360^\circ$ video generation that requires high-quality outpainting in panoramic distribution and continuously changing video masks from varying elevations. In contrast, our~\name~handles the perspective-to-$360^\circ$ video generation with global and local constraints from dual-branch diffusion and elevation-aware designs to handle changing video masks, producing high-quality $360^\circ$ video generations from perspective video anchors.

%-------------------------------------------------------------------------
\subsection{360 Panorama Generation}
Early methods~\cite{wang2022stylelight,chen2022text2light,omnidreamer,inout,boundless,bips,DBLP:journals/tmm/WuTJZQSY23} exploit GAN-based framework for panorama image generation. 
OmniDreamer~\cite{omnidreamer} proposes transformer-based framework for 360-degree outpainting and devises circular inference to obtain $360^\circ$ close-loop continuity. 
Recently, diffusion-based methods~\cite{panodiff,panogen,wu2024panodiffusion,panfusion,stitchdiffusion,feng2023diffusion360,lee2023syncdiffusion,stan2023ldm3d,zhang2024taming} have dominated image-level panorama generation. 
Due to the data scarcity of large panorama image datasets, methods~\cite{panodiff,stitchdiffusion} that directly fine-tune LDM to generate the panorama results in low-quality images with simple structures and sparse assets. 
PanFusion~\cite{zhang2024taming} introduces the dual-branch structure, consisting of a panorama and a perspective branch to leverage the synergy from both global and local {constraints} for high-quality text-to-panorama generation. 

Despite numerous efforts in image-level panorama generation, there are few works~\cite{360dvd,4k4dgen,vidpanos} focusing on panorama video generation. 360DVD~\cite{360dvd} takes text prompts and additional panorama video optical flow as guidance and learns a 360-Adapter on standard T2V models to generate plausible $360^\circ$ videos. 
4K4DGen~\cite{4k4dgen} animates a static panoramic image at user-selected regions with I2V pre-trained prior in a training-free manner. 
These fine-tuning-based or training-free approaches struggle to bridge the distribution gap between panoramic and perspective videos, resulting in simple natural perturbations, such as clouds moving and water running. In contrast, our~\name~benefits from the dual-branch video denoising structure with antipodal relation modeling and elevation-aware designs, resulting in more dynamic $360^\circ$ videos with rich and structured motions. 
\section{Our Approach}
\begin{figure*}[t]
	\centering
	\includegraphics[width=\linewidth]{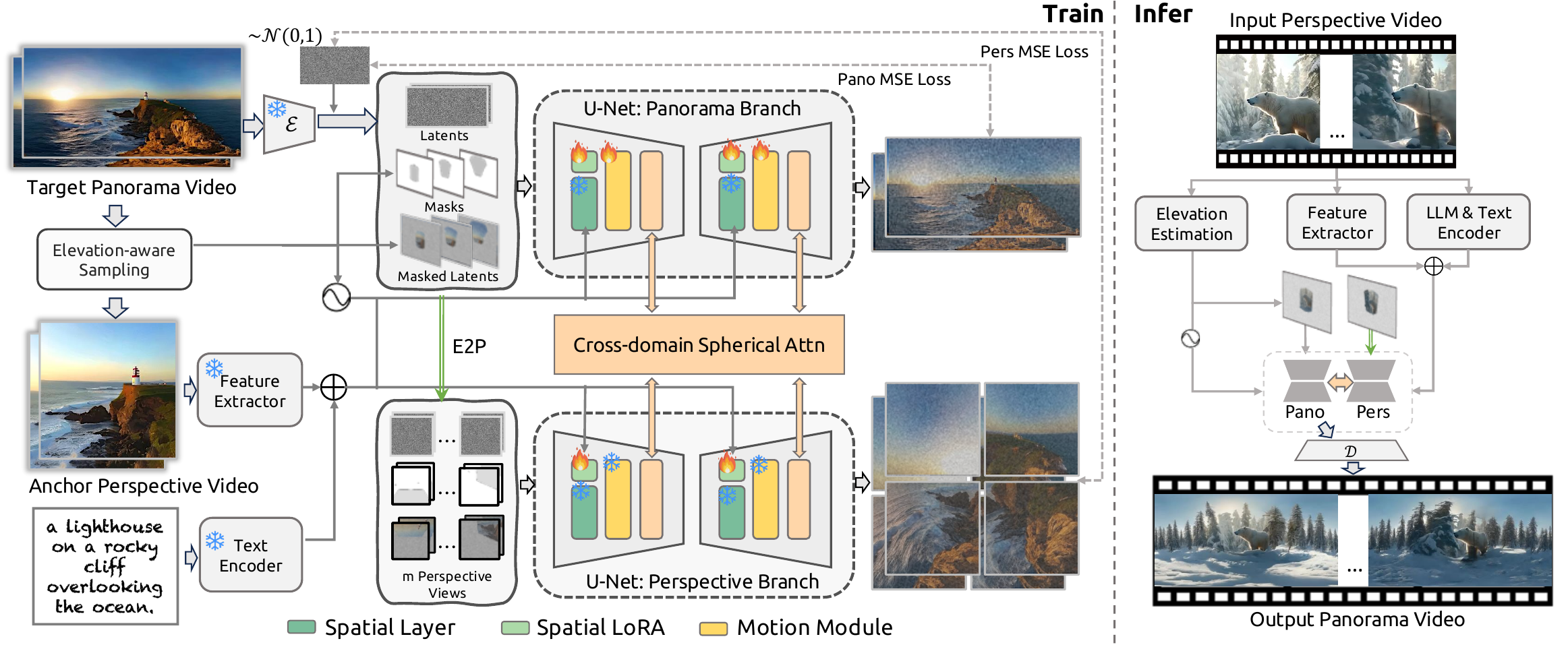}
	\caption{\small \textbf{Pipeline of~\name.} Given perspective anchor video guidance,~\name~leverages a dual-branch video noising structure, with parallelled panorama and perspective branches to denoise 360° videos with plausible panoramic patterns. Additionally, we devise the cross-domain spherical attention with antipodal masking to capture long-range dependencies for reversed antipodal motion. Finally, we introduce elevation-aware designs to handle diverse video inputs of changing elevations. }
	\label{fig:pipeline}
    \vspace{-10pt}
\end{figure*}

Given a perspective video, we aim to generate an equirectangular panoramic (ERP) video that replicates its visual appearance and motion, and extends to the complete $360^\circ \times 180^\circ$ field of view.

We propose~\name, the first perspective-to-$360^\circ$ video generation framework, featuring three key designs to create high-quality $360^\circ$ videos from a perspective video anchor, as illustrated in \cref{fig:pipeline}.
First, to learn the spherical visual and motion patterns based on pre-trained perspective generative prior, we employ the dual-branch design, consisting of a panorama branch and a perspective branch to jointly denoise the $360^\circ$ spacetime latent.
Second, to obtain more fine-grained and plausible panoramic motion, we refine the cross-domain attention to highlight antipodal masking that captures long-range motion dependencies in antipodal directions.
Finally, to handle diverse video inputs with varying elevation angles, we propose elevation-aware training and inference designs to obtain robust generations from different video anchors.

\subsection{Workflow Setup}
\noindent\textbf{Perspective-to-360° video mask construction.}
Similar to video outpainting, our task also requires a video masking on the 360° video canvas. 
For a given perspective video of $T$ frames and its camera pose $(FOV, \theta^{1:T}, \phi^{1:T})$, where $\theta$ denotes the yaw angle (azimuth) and $\phi$ denotes the pitch angle (elevation), we project it into the equirectangular domain via P2E projection to construct the $360^\circ$ video mask $M_{pano}$. By stacking the frame masks, we have the video masking $M_{pano}^{1:T}$. With this projection, we obtain the masks and masked latents to feed into the panorama UNet along with the latents.

\noindent\textbf{Anchor video preparation.}
Our training and inference have slightly different procedures to obtain the anchor videos.
During training, given the target panorama video with a set of camera poses of different elevations, we can obtain the corresponding video projections in the perspective domain and obtain a masked panorama video indicating where to outpaint and where to keep.
The anchor video for perspective domain $I_{P,anc}^{1:T}$ is the perspective video projections and the anchor video for panormama domain $I_{E,anc}^{1:T}$ is the maximum inscribed rectangle crop of the kept region in the target panorama video. During inference, we use the input perspective video as the perspective anchor, generate its text prompt with video LLM, and estimate the corresponding camera pose for mask construction and build the panorama anchor video as in training.

\noindent\textbf{Video-conditioned generation.}
The perspective-to-$360^\circ$ video task requires precise guidance over the newly generated pixels, as the unknown pixels far outnumber the given ones. To prepare the guidance signals once had anchor videos and text prompts, following~\cite{followyourcanvas}, we use SAM's visual encoder~\cite{segmentanything} to extract features from the anchor video. These features are processed by a query-based Transformer, which distills high-level visual and motion cues into compact latent tokens. To guide the massive spacetime pixel generation, we adopt the IP cross-attention from~\cite{ipadapter}, to decouple cross-attention for text prompts and visual embeddings in U-Net layers. This allows our model to extend visual content and motion patterns effectively to other views and unmasked areas of the spacetime panorama.

\subsection{Dual-Branch 360° Video Denoising}
\noindent\textbf{Dual-branch design.}
Extending a perspective video to a 360° canvas requires careful model design due to the large differences between panorama and perspective distributions. Mainstream methods either train a Latent Diffusion Model (LDM) to directly denoise panorama frames or project panorama frames to multiple perspective views for joint denoising. The former results in plain visuals and mild motion from limited panorama data, while the latter easily produces local, short-range motions in each view.

Inspired by the dual-branch design for image-level panorama generation~\cite{panfusion}, we introduce a dual-branch video denoising structure. It consists of a global panorama branch and a local perspective branch, both using U-Nets based on AnimateDiff~\cite{animatediff}, where each U-Net layer consists of spatial layers initialized from SD weights and a motion module initialized from~\cite{followyourcanvas} weights. 
Aiming to generate a $360^\circ$ panorama video of shape $\mathcal{R}^{T\times 3 \times H \times W}$ ($W = 2H$), the noisy latent input is of shape $\mathcal{R}^{T\times 9 \times H \times W}$ from latents, binary masks and masked latents. The panorama and perspective branch share the same noisy latents such that they could be aligned in the subsequent denoising steps. The panorama branch directly takes the noisy global latents as input, whereas the perspective branch projects the latent into $m=20$ perspective views according to the icosahedron modeling~\cite{rey2022360monodepth,panfusion}, and each view gets latent of shape $\mathcal{R}^{T\times 9 \times H/2 \times H/2}$ through projection. 
LoRA layers are employed at spatial self-attention and IP cross-attention on each branch to cater to different resolutions and the new panorama distribution.

Denoising on the same noisy latents, the global panorama branch provides a holistic understanding of the $360^\circ$ spacetime canvas to maintain the global consistency, while the local perspective branch leverages good pre-trained perspective generative prior to preserve rich local details at each view in every frame. At the end of each U-Net block, the latents from both branches are aligned via a cross-domain attention module to enhance the local-global synergy.
The attention module works bi-directionally. Panorama pixels and $m$ perspective window pixels from the same frame are flattened into two sequences and feed into the attention module as the key-value token or query token in alternation, with spherical positional encoding~\cite{panfusion} added to indicate the relative spherical location. The information exchange is highlighted for directly mapped pixel pairs between two domains. 
Specifically, as illustrated in \cref{fig:method_antipodal}, at each frame, we use P2E projections to map the pixels from the panorama branch to the directly mapped pixels (\textcolor{orange}{orange}) in the perspective branch, and vice versa. This bidirectional mapping is reflected via an attention mask, denoted spherical mask, and added to the attention module. Gaussian blur is also applied on the spherical mask to enable neighboring activations in the attention.

\noindent\textbf{Resource-friendly fine-tune strategy.} 
With only thousands of training panorama videos, we seek to fine-tune a limited proportion of parameters to yield good generation results. Thanks to the space-time disentangled design from AnimateDiff~\cite{animatediff}, we do not need to fine-tune the whole model to learn the distributions of panoramic videos. In practice, we add LoRA on spatial attention layers for both branches. To learn panoramic motion, we show in \cref{supp:finetune} that employing motion LoRA layers does not have sufficient influence on the pretrained prior to generate good panorama motion.
Therefore, the whole motion module is fine-tuned for panorama branch to learn the necessary prior for generating spherical motion patterns. 
Note that, to fully exploit the perspective generative prior and prevent our limited panorama video data messing up the good perspective motion from pretrained weights, we do not fine-tune the motion module in the perspective branch. Experimental results show that our resource-friendly fine-tuning strategy achieves stunning results for $360^\circ$ video generation despite the limited data and GPU resources. The training objectives of the trainable modules in the two branches are:
\begin{equation}
\small
\resizebox{.9\hsize}{!}{$\mathcal{L}_{E}=\mathbb{E}_{\mathcal{E}\left(x_{E}^{1: T}\right), y, \epsilon_{E}^{1: T}, t}\left[\left\|\epsilon_{E}  - \epsilon_{E,\Theta}\left(z_{E,t}^{1: T}, t, \tau_{\Theta}(y),I_{E,anc}^{1:T}\right)\right\|_{2}^{2}\right] ,$}
\end{equation}
\begin{equation}
\small
\resizebox{.9\hsize}{!}{$\mathcal{L}_{P}=\mathbb{P}_{\mathcal{P}\left(x_{P}^{1: T}\right), y, \epsilon_{P}^{1: T}, t}\left[\left\|\epsilon_{P}  - \epsilon_{P,\Theta}\left(z_{P,t}^{1: T}, t, \tau_{\Theta}(y) ,I_{P,anc}^{1:T}\right)\right\|_{2}^{2}\right],$}
\end{equation}
where $x_{E}^{1: T}$ and $\epsilon_{E,\Theta}$ represent the target panorama video and the predicted noise from the panorama branch, respectively, while $x_{P}^{1: T}$ and $\epsilon_{P,\Theta}$ denote the target perspective video and the predicted noise from the perspective branch. To facilitate the synchronous exchange of information between two branches, we combine the two branches losses as the final training objective $\mathcal{L}=\mathcal{L}_{E} + \frac{1}{m} \sum_{i=1}^{m} \mathcal{L}_{P}^{i}$.

\subsection{Encouraging Fine-grained Panoramic Patterns}
The dual-branch design generates plausible spherical patterns for spatial appearance and motion, however, to achieve fine-grained panoramic videos, i.e. $360^\circ$ close-loop continuity and reversed antipodal motion, we introduce additional advanced techniques, including circular padding and antipodal mask in cross-domain spherical mask.

\noindent\textbf{Circular padding for close-loop continuity.}
The $360^\circ$ close-loop continuity is an important property for panorama media, which refers to the left-most and right-most edges of the panorama video seamlessly connected to form a continuous loop. Following~\cite{panfusion,layerpano3d}, we employ the circular padding before the convolutional layer in each U-Net block and unpad to its original resolution afterward to mitigate the loop inconsistency from local convolutions. For a better 360 immersive experience, in addition to the main framework, we also implement circular padding in the optional video super-resolution post-processing~\cite{venhancer} to pad and unpad the upsampled latents in the decoding process for SR close-loop continuity.

\begin{figure}[t]
	\centering
	\includegraphics[width=\linewidth]{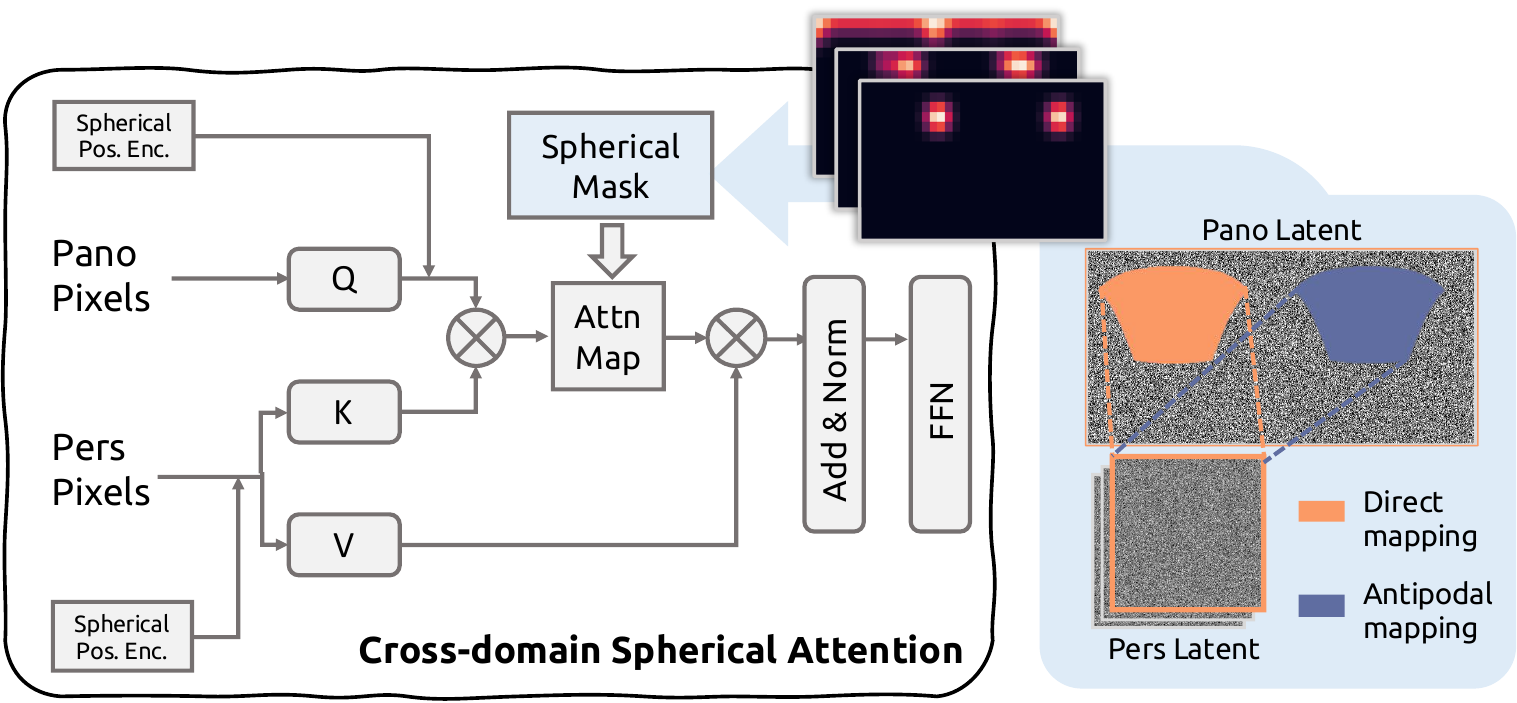}
	\caption{\small \textbf{Cross-domain Spherical Attention} (perspective branch) highlights interaction for direct-mapped pixels (spherical mask) and \textbf{antipodal-mapped} pixels (antipodal mask) between panorama and perspective domains.}
	\label{fig:method_antipodal}
    \vspace{-20pt}
\end{figure}

\noindent\textbf{Antipodal mask for reversed motion.}
One particular property of panoramic motion is that the antipodal pixel always exhibits the reversed camera motion. That is to say, if the camera is moving forward in the forward viewing direction, then if the viewer turns to the backward viewing direction, the scene should look like moving away from the viewer. In both the panoramic branch and the perspective branch, the antipodal pixels are usually distant from each other in the token sequence, whereas the attention modules often tend to attend more on neighboring pixels, we observe that the reversed motion in the antipodal directions is mild compared to the motion in the neighboring pixels of the projected guidance video. To emphasize the antipodal relations in the $360^\circ$ videos, we highlight the antipodal activations in the cross-domain spherical mask and denote this set of activations as the \textit{antipodal mask}. As shown in \cref{fig:method_antipodal}, we find each pixel in the perspective domain the antipodal pixel ({\color{BlueViolet}blue}) of its panorama counterpart, and vice versa. The antipodal pairs are assigned high activation on the masking and added to the attention map. Gaussian blur is also applied on the mask for antipodal neighbors. 

\subsection{Elevation-robustness over General Video Input}
\begin{figure}[t]
	\centering
	\includegraphics[width=0.98\linewidth]{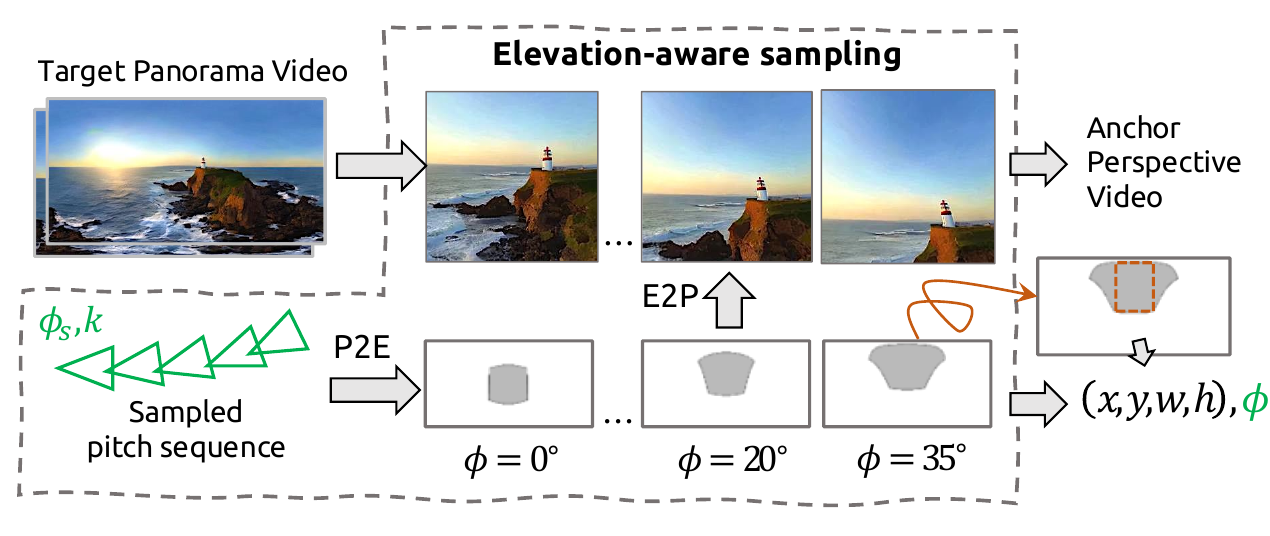}
    \vspace{-15pt}
	\caption{\small \textbf{Elevation-aware sampling} augments the training samples with diverse elevation trajectories.}
	\label{fig:method_elevation}
    \vspace{-15pt}
\end{figure}

Commonly, we assume the video input to be upright with a fixed camera pose, but in-the-wild perspective videos are of various azimuthal angles $\theta^{1:T}$ and elevation angles $\phi^{1:T}$. As the ERP panorama exhibits $360^\circ$ close-loop continuity, without loss of generality, we can set the azimuthal angle to zero by assuming the viewer is rotating with the camera to simplify computation. However, we must specifically investigate $\phi$ as the change of elevation angle significantly impacts the shape and location of the masking, as shown in~\cref{fig:method_elevation}. To create an elevation-augmented generation pipeline, we propose elevation-augmented data sampling, an elevation estimator for inference.

\noindent\textbf{Elevation-aware data sampling.} 
We create flexible masking for each frame with an elevation-aware data sampling strategy. As illustrated in \cref{fig:method_elevation}, for each sample, we randomly sample $\phi_s \in (-20^\circ,20^\circ)$ and a slope ratio $k \in (-0.5, 0.5)$, then construct a smooth pitch trajectory sequence with $\phi_s$ as the starting angle and assign frame $t\in (1,T)$ with $\phi = \phi_s + k\cdot t$. 
As the masks exhibit non-gridded structure from the spherical P2E projection, we compute the positional encoding by calculating the maximum inscribed rectangle of each mask, shown with a red dashed box. The center coordinates $x,y$, and size $h,w$ are combined with the $360^\circ$ canvas size $H, W$ to compute the sinusoidal embedding as the mask positional encoding on the panorama branch. Moreover, to highlight the pitch angle, we also compute the sinusoidal embedding for $\theta$ and concatenate with the mask positional encoding as the final positional encoding.

\noindent\textbf{Elevation estimation for inference.}
During inference, to construct a pitch sequence for each test video, we employ PerspectiveFields~\cite{perspectivefields} to estimate the pitch angle for each input frame. As PerspectiveFields is developed for single image calibration, estimating the pitch sequence frame by frame causes inconsistency in the projected mask sequence. Therefore, we leverage LinearRegression to fit the pitch sequence and produce smooth masking on the $360^\circ$ video canvas.

\subsection{Training Data Collection}
Due to the significant domain gap between panoramic and perspective motions, fine-tuning requires extensive text-video paired data in the panoramic domain. Although the WEB360~\cite{360dvd} dataset offers 2,114 panoramic videos of outdoor landscapes, it remains limited in scale and lacks diverse, structured motions typical in urban environments, such as vehicle traffic and pedestrian activities. To address this, we collected 8,630 360-degree videos from YouTube, covering virtual city tours, wildlife documentaries, and VR game captures, to supplement training. The videos were converted to equirectangular format, segmented into consecutive shots using TransNet-v2~\cite{transnetv2}, and processed into 5-second clips at 20 fps with a 2× speed-up. We extracted optical flow using PanoFlow~\cite{panoflow}, to filter out static clips based on maximum optical flow values. Text captions were generated as two-sentence summaries for each clip using VideoLLaMa-2~\cite{videollama2}. We conduct our training on a combination of the WEB360 and our dataset with 10,744 samples.

\begin{figure*}[t]
	\centering
	\includegraphics[width=0.98\linewidth]{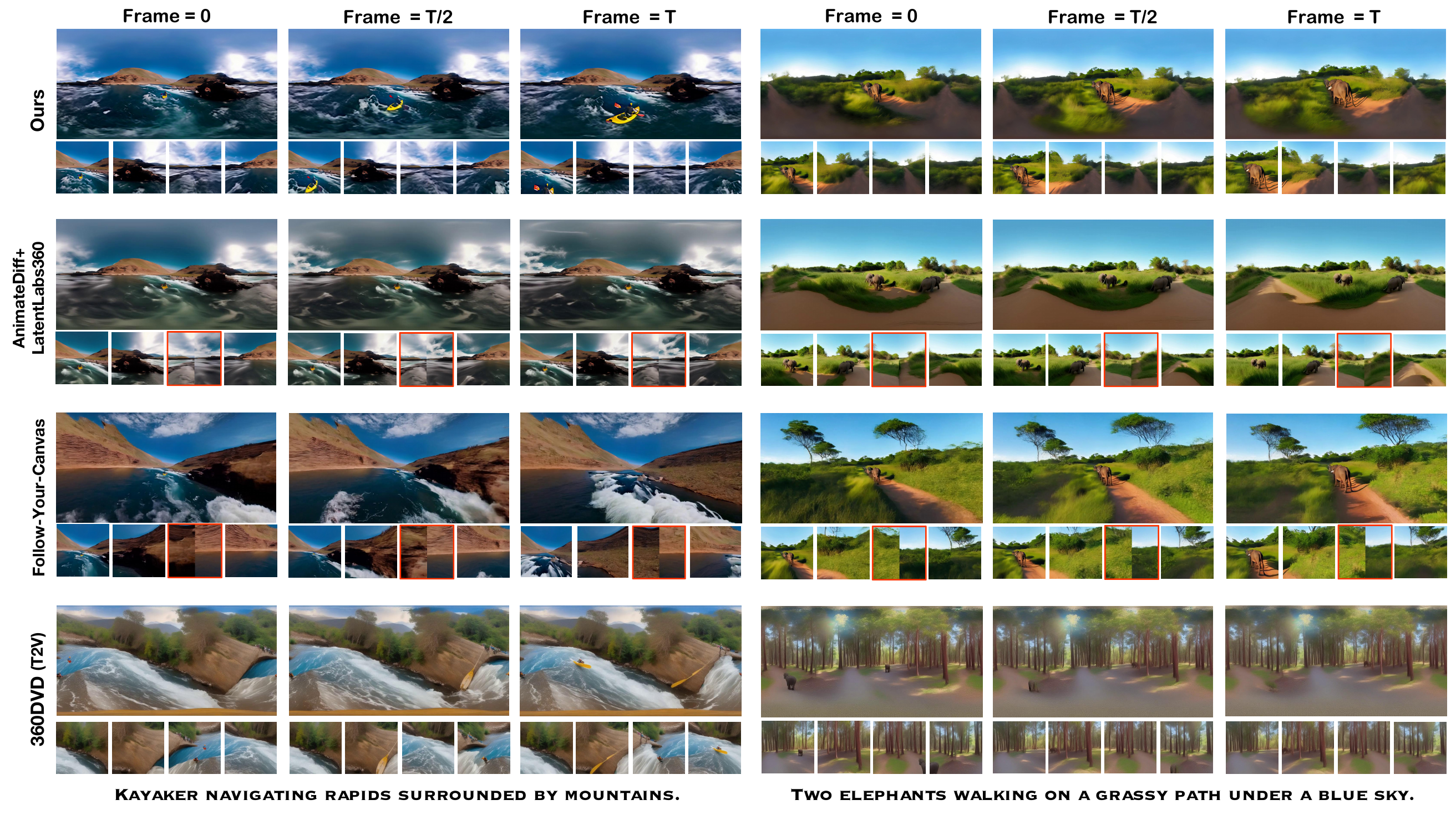}
        \vspace{-10pt}
	\caption{\small \textbf{Qualitative comparisons} on 360° video generations among state-of-the-art methods.~\name~generates 360° video generation with superior visual quality and plausible panoramic patterns.}
	\label{fig:qualitative_video}
\end{figure*}

\begin{table*}[ht]
    \small
    \centering
    \setlength{\tabcolsep}{5pt}
    \renewcommand{\arraystretch}{0.9}
    \vspace{-6pt}
    \begin{tabular}{@{}l|ccccc@{}}
    \toprule
    Method & Imaging Quality~$\uparrow$ & Aesthetic Quality~$\uparrow$ & Motion Smoothness~$\uparrow$ & Subject Consistency~$\uparrow$ & VQA~$\uparrow$ \\ 
    \midrule
        Animatediff~\cite{animatediff}+LatentLab360  & 0.7159 & 0.5840 & 0.9780 & 0.9338  & 0.7860\\ 
    Follow-Your-Canvas~\cite{followyourcanvas}  & 0.6978 & 0.6432  & 0.9771 & 0.9529  & 0.8444\\ 
        360DVD~\cite{360dvd} & 0.5537 & 0.4745  & 0.9701 & 0.9629  & 0.5867 \\        
        Ours  & \textbf{0.7487} & \textbf{0.6439} & \textbf{0.9806} & \textbf{0.9710}  & \textbf{0.8672} \\
        \bottomrule
    \end{tabular}
    \vspace{-8pt}
    \caption{\textbf{Quantitative comparison} with SoTA approaches on Vbench~\cite{vbench} metrics and VQA from Q-Align~\cite{qalign}.}
    \vspace{-8pt}
  \label{tab:video_metrics}
\end{table*}

\begin{table*}[t]
    \small
    \centering
    \setlength{\tabcolsep}{8pt}
    \renewcommand{\arraystretch}{0.9}
    \vspace{-6pt}
    \scalebox{0.98}{\begin{tabular}{@{}l|ccccc@{}}
    \toprule
    Method & Imaging Quality~$\uparrow$ & Aesthetic Quality~$\uparrow$ & Motion Smoothness~$\uparrow$ & Subject Consistency~$\uparrow$ & VQA~$\uparrow$ \\ 
    \midrule
        Ours(Full model)  & \textbf{0.7487} & \textbf{0.6439} & \textbf{0.9806} & \textbf{0.9710}  & \textbf{0.8672} \\
        w/o persbranch   & 0.7050 & 0.6124 & 0.9744 & 0.9599  & 0.8435\\ 
        w/o panobranch  & 0.5854 & 0.4549  & 0.9700 & 0.9546  & 0.6168\\ 
        w/o antipodal mask & 0.7446 & 0.6360  & 0.9802 & 0.9652  & 0.8580 \\         
        w/o elevation-aware designs & 0.6978 & 0.6296  & 0.9771 & 0.9529  & 0.7663\\        
        \bottomrule
    \end{tabular}}
     \vspace{-8pt}
     \caption{\textbf{Quantitative ablation studies} on dual-branch designs, antipodal mask, and elevation-aware designs.}
    \vspace{-2pt}
  \label{tab:ablation}
\end{table*}

\begin{table}[t]
    \small
    \centering
    \vspace{-6pt}
\scalebox{0.85}{\begin{tabular}{l|ccc}
\toprule
Method                   & \multicolumn{1}{l}{\begin{tabular}[c]{@{}l@{}}Graphics \\ Quality~$\uparrow$\end{tabular}} & \multicolumn{1}{l}{\begin{tabular}[c]{@{}l@{}}Structure \\ Plausibility~$\uparrow$\end{tabular}} & \multicolumn{1}{l}{\begin{tabular}[c]{@{}l@{}}Temporal \\ Coherence~$\uparrow$\end{tabular}} \\ \midrule
Animatediff~\cite{animatediff}+LatentLab360 & 2.3488& 2.3894& 2.1433\\
360DVD~\cite{360dvd}& 1.2067& 1.7588& 1.4279\\
Follow-Your-Canvas~\cite{followyourcanvas}& 2.7692& 2.1298& 3.0385\\ 
Ours & \textbf{3.6827} & \textbf{3.7260} & \textbf{3.3942} \\ \bottomrule
\end{tabular}}
    \vspace{-8pt}
\caption{\textbf{Human evaluation results} on $360^\circ$ video generation. }
    \vspace{-2pt}
  \label{tab:userstudy}
\end{table}

\section{Experiments}
\subsection{Implementation Details}
\noindent\textbf{Training settings.}
The spatial and motion modules in our model initializations are respectively based on Stable Diffusion v2.1 and Follow-Your-Canvas~\cite{followyourcanvas}. Training is conducted on 8 NVIDIA A100 GPUSs with 50k training steps using our proposed dataset, with the spatial LoRA rank and $ \alpha_{LoRA}$ set to 32 and 1.0. The resolution is set to $512 \times 1024$, the length of frames to 40, the batch size to 1, and the learning rate to $1 \times 10^{-5}$.

\noindent\textbf{Evaluation metrics.}
We employ video quality assessment (VQA) metric from Q-Align~\cite{qalign} to measure the general quality of $360^\circ$ videos. In addition, we project the $360^\circ$ videos into 4 perspective views with $\phi=0^\circ, FOV=90^\circ,\theta=[0^\circ,90^\circ,180^\circ,270^\circ]$ and employ imaging quality, aesthetic quality, motion smoothness, and subject consistency metrics from Vbench~\cite{vbench} to measure the graphics quality and motion consistency of projected perspective videos as well as examine the panoramic structure plausibility of the $360^\circ$ videos.

\noindent\textbf{Comparison methods.}
Since we are the first perspective-to-$360^\circ$ video generation framework, it's infeasible to find a method that has the exactly same input condition as ours. We compare with methods that produces 360 videos from other kinds of conditions.
\textbf{360DVD}~\cite{360dvd} is the advanced text-guided $360^\circ$ video generation method, that takes text prompts and panorama video optical flow as input for $360^\circ$ video generation. Here, we feed the same text prompt as ours with its default optical flow into 360DVD for comparison.
\textbf{Follow-Your-Canvas}~\cite{followyourcanvas} is the state-of-the-art video outpainting method that employs tile-based outpainting to handle outpainting of arbitrary size and resolution. Here, we project our perspective anchor video into the panorama canvas and feed the masked canvas into~\cite{followyourcanvas} for evaluation. 
\textbf{AnimateDiff+LatentLab360}~\cite{animatediff} takes image inputs and animates with the LatentLab360 LoRA. For comparison, we feed the first frame of our generations into AnimateDiff+LatentLab360 and take its animated videos for evaluation.

\subsection{Quantitative Comparisons}
\cref{tab:video_metrics} provides the quantitative comparison of~\name~with other models on a total of 100 test cases.~\name~achieves the best performance in all metrics. This indicates that our approach excels other methods not only in general panorama video quality but also in panorama structure plausibility. We also involve human evaluation to examine the quality of generated $360^\circ$ videos. We invite 26 users with expertise in video and 3D generation to assess the results across three dimensions: graphics quality, structure plausibility, and temporal coherence. \cref{tab:userstudy} reports the average user ranking of all four methods, and our method achieves the best performance in all three dimensions. Please refer to \cref{supp:implement_detail} for more details regarding the setup and metrics. Moreover, we show that our method also achieves superior panorama image outpainting performance. Please refer to \cref{supp:pano_img_outpaint} for comparison.

\subsection{Qualitative Comparisons}
We present the qualitative comparison between~\name~and other SoTA models in \cref{fig:qualitative_video}. 
At each frame, we present the panorama frame with four projected perspective views ($\phi=0^\circ, \theta=[0^\circ,90^\circ,180^\circ,270^\circ]$) to make it easier to examine the panoramic structure plausibility. \cref{fig:qualitative_video} shows that AnimateDiff+LatentLab360 and Follow-Your-Canvas fail to achieve $360^\circ$ close-loop continuity (orange box) and produce mild-scale motion by observing the change of the canoe location (left-side) and animal size (right-side). 360DVD produces more distorted patterns and blurry visual appearance in both cases. In contrast, our~\name~achieves superior visual quality and plausible, obvious motion in the generated $360^\circ$ videos. We highly recommend readers visit our project page for more stunning results.

\begin{figure}[t]
	\centering
	\includegraphics[width=\linewidth]{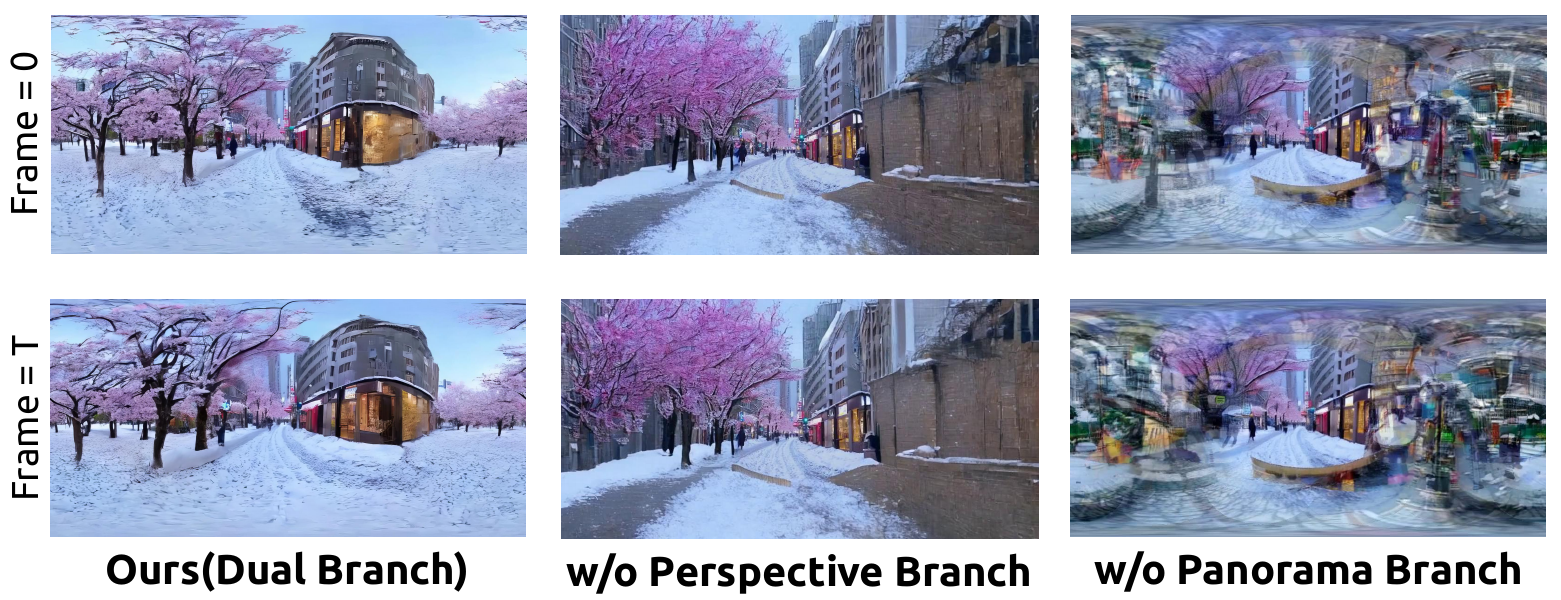}
        \vspace{-5pt}
	\caption{\small \textbf{Qualitative ablation on the dual-branch design} shows that dual-branch design generates plausible panoramic patterns compared to single-branch. }
	\label{fig:ablation_dualbranch}
    \vspace{-5pt}
\end{figure}

\begin{figure}[t]
	\centering
	\includegraphics[width=\linewidth]{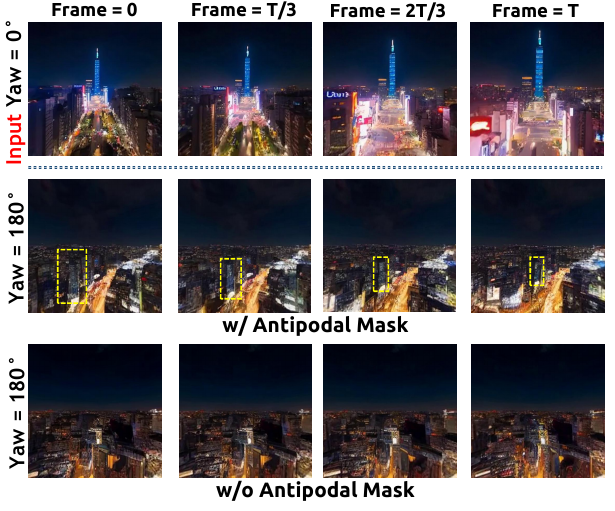}
        \vspace{-10pt}
	\caption{\small \textbf{Qualitative ablation on the antipodal mask} shows improved reversed motion in the antipodal view from the antipodal activations.}
	\label{fig:ablation_antipodal}
    \vspace{-10pt}
\end{figure}

\begin{figure}[t]
	\centering
	\includegraphics[width=\linewidth]{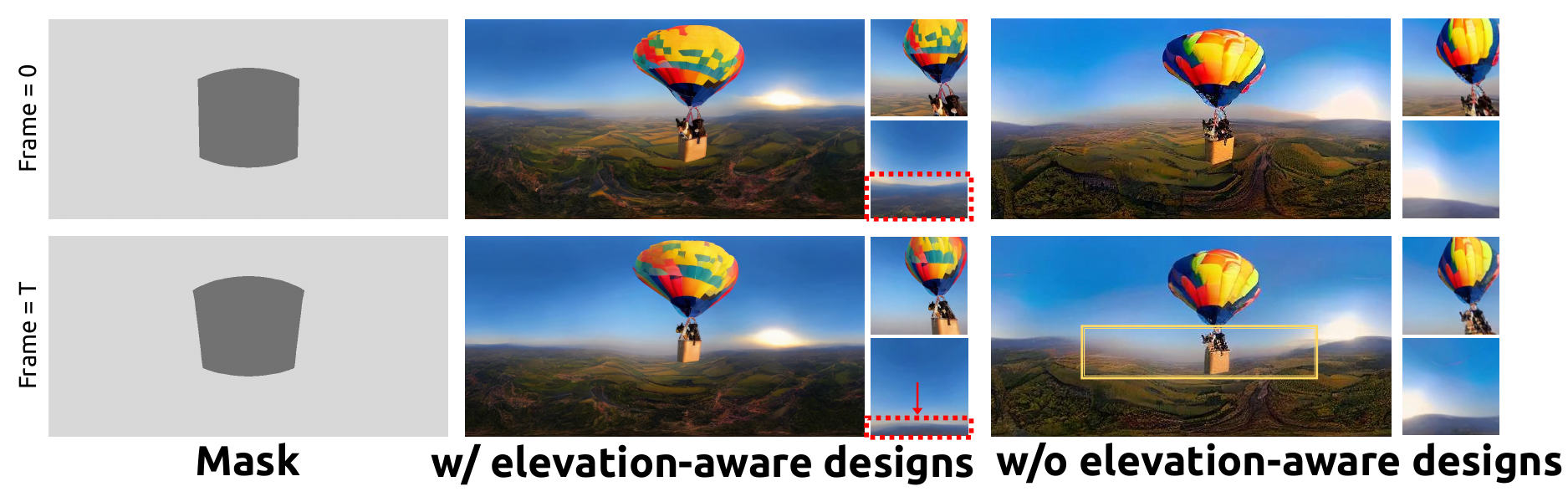}
	\caption{\small \textbf{Qualitative ablation on the elevation-aware designs} shows reduced artifacts from the elevation-aware designs .}
	\label{fig:ablation_elevation}
    \vspace{-10pt}
\end{figure}

\subsection{Ablative Studies}
\noindent\textbf{Ablation on dual-branch design.} We show the qualitative results of our model with dual-branch design (full model), with single panorama branch and single perspective branch in \cref{fig:ablation_dualbranch}. Either single panorama branch or single perspective branch alone is insufficient to achieve good spherical visual structure and panoramic motion patterns with limited high-quality $360^\circ$ videos. Quantitative results in \cref{tab:ablation} also demonstrate the effectiveness of the proposed dual-branch video denoising branch. 
Further ablation on dual-branch fine-tuning can be found in \cref{supp:finetune}.

\noindent\textbf{Ablation on antipodal mask.} \cref{fig:ablation_antipodal} illustrates the impact of the antipodal mask on $360^\circ$ video generations. We show the perspective videos of $\theta=0^\circ$ and $\theta=180^\circ$ at frame $0, T/3, 2T/3, T$. Thanks to the antipodal mask, as the camera moves forward in the input direction  ($\theta=0^\circ)$, we can observe clear backward motion in the antipodal views ($\theta=180^\circ)$. Quantitative result in \cref{tab:ablation} also proves the effectiveness of adding antipodal activations in the cross-domain spherical mask.

\noindent\textbf{Ablation on elevation-aware designs.} 
In \cref{fig:ablation_elevation}, we show that not addressing changing elevations in the input anchor video leads to artifacts in the $360^\circ$ video's scene geometry. Without elevation-aware handling, the generated video fails to cater to the elevation in the input anchor video, causing distortion in scene elements like mountains (yellow box). Quantitative results in \cref{tab:ablation} further validates the benefits of elevation-aware designs.

\section{Conclusion}
In this paper, we propose~\name, the first perspective-to-360° video generation framework using video-based control to create $360^\circ$ videos with rich and structured panoramic motions. Our contributions are three-fold: (1) we introduce a dual-branch video denoising structure with panorama and perspective branches for $360^\circ$ video generation with both global and local constraints; (2) we design an antipodal mask in cross-domain spherical attention to model long-range motion dependencies in panoramic videos; and (3) elevation-aware designs are introduced to handle varying elevation angles in diverse video inputs. Experiments show that~\name~produces 360° videos with superior video quality and panoramic motion plausibility.

\noindent\textbf{Limitations and future works.}
~\name~leverages off-the-shelf models to estimate elevation for inference, therefore the generated video could contain artifacts from inaccurate elevation estimations. In the next step, we plan to train a specialized elevation estimation module with our $360^\circ$ video dataset to improve the estimation accuracy. 

{
    \small
    \bibliographystyle{ieeenat_fullname}
    \bibliography{main}
}

\renewcommand\thesection{\Alph{section}}
\renewcommand\thefigure{\Alph{figure}}
\renewcommand\thetable{\Alph{table}}

\setcounter{section}{0}
\setcounter{table}{0}
\setcounter{figure}{0}

\section{Discussion on Panorama Image Outpaint}
\label{supp:pano_img_outpaint}
A bonus advantange of~\name~is that apart from panorama video outpainting, we also achieves superior performance for panorama image outpainting. We compared our method with state-of-the-art panorama image outpainting approaches, including Diffusion360~\cite{feng2023diffusion360}, PanoDiffusion~\cite{wu2024panodiffusion} and SIG-SS~\cite{hara2022sig-ss}. We use the first frame of a video as the input image and extract the first frame of our outpainted video as our result for panorama image outpainting. For quantitative comparison, we adopt \textbf{Intra-Style}~\cite{intra-style,lee2023syncdiffusion} metric to evaluate the panorama style coherence; \textbf{CLIP}~\cite{Clipscore} to measure the alignment between the panorama and the input text prompts; \textbf{IQA}~\cite{qalign} measures the overall image quality. \cref{tab:image_comparison} shows our method achieves the best performance among the compared methods across all metrics.

\begin{figure*}[t]
	\centering
	\includegraphics[width=\linewidth]{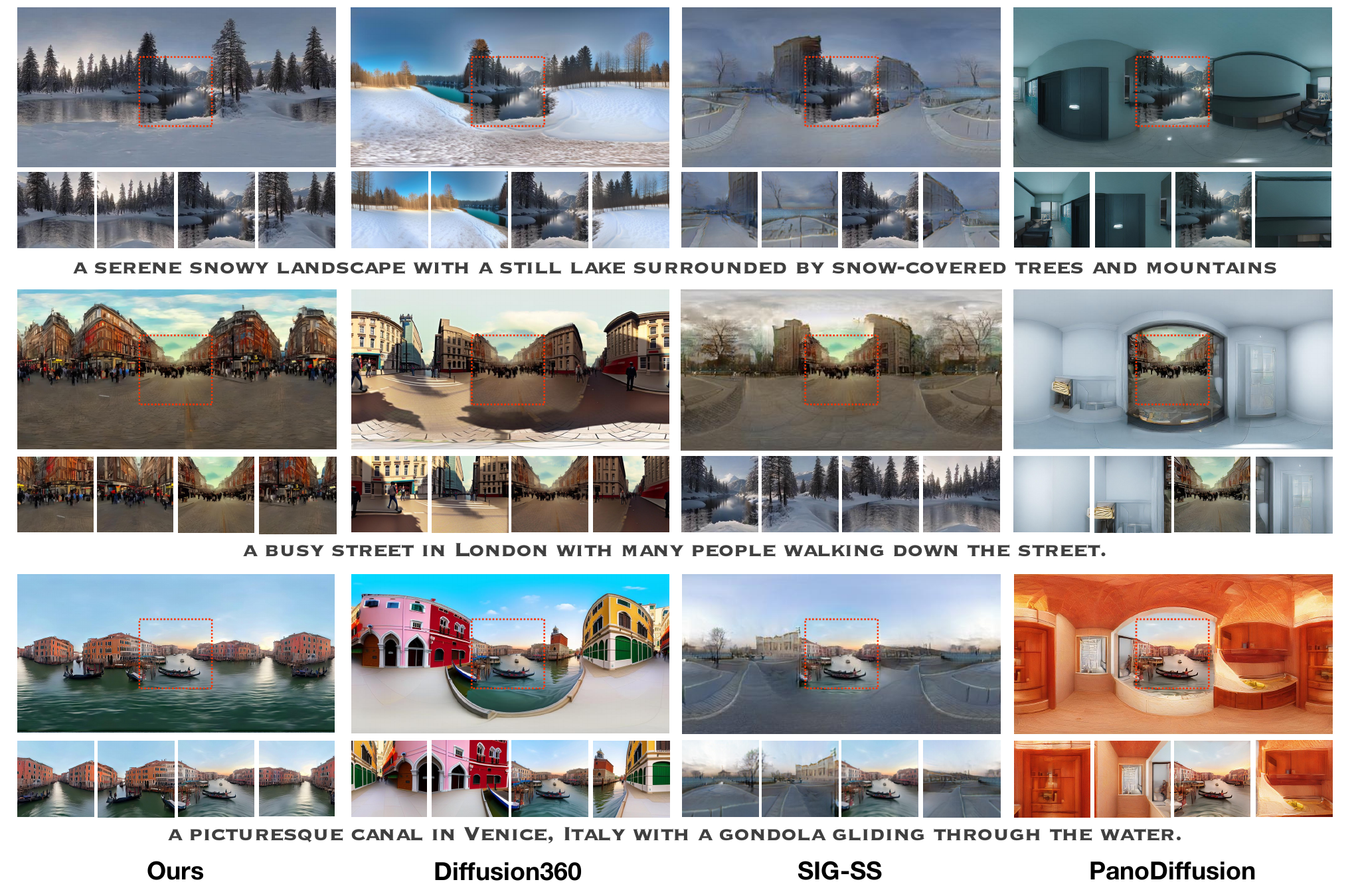}
	\caption{\small \textbf{Qualitative comparisons between~\name~and the state-of-the-art panorama outpainting methods.} }
	\label{fig:image_outpaint}
    % \vspace{-10pt}
\end{figure*}

We also show the qualitative comparison in \cref{fig:image_outpaint}, with red dashed box indicating the input image. The results of Diffusion360~\cite{feng2023diffusion360} show less consistency that its newly generated pixels having different style with the know pixels. SIG-SS~\cite{hara2022sig-ss} is a Gan-based methods and its generations exhibits over-smoothness compared diffusion-based approaches. PanoDiffusion~\cite{wu2024panodiffusion} focuses on indoor scenes and does not generalize well to outdoor scene. In contrast,~\name~outpaints more consistent, high-quality and aesthetic panorama compared to other approaches.

\begin{table}[ht]
    % \scriptsize
    \small
    \centering
    
    \vspace{-6pt}
    \setlength{\tabcolsep}{7pt}
    \begin{tabular}{@{}l|cccc@{}}
    \toprule
    Method & Intra-Style($\times 10^{-3}$)~$\downarrow$ & CLIP~$\uparrow$ & IQA~$\uparrow$  \\ 
    \midrule
        Diffusion360~\cite{feng2023diffusion360} & 3.40 & 27.49 & 0.77 \\ 
        PanoDiffusion~\cite{wu2024panodiffusion}  & 3.46  & 23.19 &  0.72 \\ 
            SIG-SS~\cite{hara2022sig-ss} &  2.06 & 26.26 & 0.48 \\ 
        Ours & \textbf{0.99} & \textbf{29.12} & \textbf{0.78}  \\
        \bottomrule
    \end{tabular}
    \vspace{-8pt}
    \caption{\textbf{Quantitative comparison with the state-of-the-art Panorama Outpainting methods.} }
  \label{tab:image_comparison}
\end{table}

\section{Extended Ablations}
\begin{table*}[t]
    \small
    \centering
    \setlength{\tabcolsep}{10pt}
    \renewcommand{\arraystretch}{0.9}
    \vspace{-6pt}
    \scalebox{0.97}{\begin{tabular}{c|ccc|ccccc}
\toprule
     & \begin{tabular}[c]{@{}c@{}}Pano \\ Spatial LoRA\end{tabular} & \begin{tabular}[c]{@{}c@{}}Pers \\ Spatial LoRA\end{tabular} & \begin{tabular}[c]{@{}c@{}}Pano \\ Motion Module\end{tabular} & IQ~$\uparrow$  & AQ~$\uparrow$  & MS~$\uparrow$  & SC~$\uparrow$  & VQA~$\uparrow$  \\ \midrule
 \#1 (ours) & \checkmark &  \checkmark  &  \checkmark  &  \textbf{0.7487}  & \textbf{0.6439} & \textbf{0.9806} & \textbf{0.9710} & \textbf{0.8672} \\
\#2 & \checkmark  & & \checkmark & 0.7474 & 0.6235  & 0.9756 & 0.9664 & 0.8584 \\ 
\#3 & & \checkmark & \checkmark & 0.7305 & 0.6246 & 0.9720 & 0.9657 & 0.8553  \\ 
\#4 & \checkmark &  \checkmark  & Motion LoRA & 0.7242 & 0.6067 & 0.9758 & 0.9558  &0.8385 \\
 \bottomrule
\end{tabular}}
     \vspace{-8pt}
     \caption{\textbf{Ablative study on dual-branch fine-tune strategy} across Image Quality (IQ), Aesthetic Quality (AQ), Motion Smoothness (MS), Subject Consistency (SC) and VQA metrics.}
    \vspace{-2pt}
  \label{tab:ablation_finetune}
\end{table*}

\subsection{Ablation on Dual-branch Fine-tuning Strategy}
\label{supp:finetune}
We ablate on the fine-tuning strategy that trains our framework to produce high-quality generations with limited data and computational resources. 
As shown in \cref{tab:ablation_finetune}, due to the gap between the domain of the pretrained model and the panorama video domain, removing the spatial LoRA layer in any branch will reduce the video quality to a certain extent. Thus, compared with directly fine-tuning all spatial modules, applying LoRA to both branches proves to be a more cost-effective choice.

For motion modeling, to effectively capture panoramic motion patterns, it is intuitive to add motion LoRA layer or finetune the motion module on the panorama branch. We further conduct a comparative experiment between directly fine-tuning motion module and applying motion LoRA. Results show that compared to the spatial domain gap, the temporal domain gap between the pretrained model and panoramic videos is more substantial. Consequently, simply utilizing LoRA for fine-tuning can result in overfitting, ultimately degrading the quality of generated video.

\subsection{Ablation on Extended Web Data}
\label{supp:webdata}
As we collect additional panorama video data from web in our training, we also ablate on the effect of the extended data. We train 360DVD~\cite{360dvd} on our training data and report the performance in \cref{tab:our_data}. Results show that the performance of 360DVD~\cite{360dvd} improves across all metrics using our data but remains weak compared to~\name, demonstrating the effectiveness of both our data and our proposed framework designs.

\subsection{Ablation on Elevation Estimation Smoothing}
\label{supp:elevation_smooth}
As demonstrated in \cref{tab:smooth}, the absence of elevation estimation smoothing leads to a noticeable decline in the quality of the generated video on each quantitative metric. 
This is because the adopted off-the-shelf elevation predictor~\cite{perspectivefields} is primarily designed for images rather than videos, and it handles videos as a sequence of images without handling the noise and temporal inconsistency between frames. Consequently, it produces inconsistent and oscillatory predictions for videos without additional smoothing. By incorporating this smoothing technique, the adverse effects of these fluctuations are mitigated to a significant extent, thereby enhancing the overall quality of the generated video.

\begin{table*}[t]
    \small
    \centering
    \setlength{\tabcolsep}{8pt}
    \renewcommand{\arraystretch}{0.9}
    \vspace{-6pt}
    \scalebox{0.98}{\begin{tabular}{@{}l|ccccc@{}}
    \toprule
    Method & Imaging Quality~$\uparrow$ & Aesthetic Quality~$\uparrow$ & Motion Smoothness~$\uparrow$ & Subject Consistency~$\uparrow$ & VQA~$\uparrow$ \\ 
    \midrule
        Ours  & \textbf{0.7487} & \textbf{0.6439} & \textbf{0.9806} & \textbf{0.9710}  & \textbf{0.8672} \\
        360DVD~\cite{360dvd} & 0.5537 & 0.4745  & 0.9701 & 0.9629  & 0.5867 \\     
        360DVD~\cite{360dvd}  + Our Data & 0.6991 & 0.5381  & 0.9739 & 0.9696  & 0.7573\\        
        \bottomrule
    \end{tabular}}
     \vspace{-8pt}
     \caption{\textbf{Ablative study on our extended panorama video dataset.}}
    \vspace{-2pt}
  \label{tab:our_data}
\end{table*}

\begin{table*}[t]
    \small
    \centering
    \setlength{\tabcolsep}{10pt}
    \renewcommand{\arraystretch}{0.9}
    \vspace{-6pt}
    \scalebox{1}{\begin{tabular}{@{}l|ccccc@{}}
    \toprule
    Method & Imaging Quality~$\uparrow$ & Aesthetic Quality~$\uparrow$ & Motion Smoothness~$\uparrow$ & Subject Consistency~$\uparrow$ & VQA~$\uparrow$ \\ 
    \midrule
        Ours  & \textbf{0.7487} & \textbf{0.6439} & \textbf{0.9806} & \textbf{0.9710}  & \textbf{0.8672} \\     
        w/o smoothing & 0.7408 & 0.6082  & 0.9706 & 0.9512  & 0.8443\\        
        \bottomrule
    \end{tabular}}
     \vspace{-8pt}
     \caption{\textbf{Ablative study on Elevation Estimation Smoothing.}}
    \vspace{-2pt}
  \label{tab:smooth}
\end{table*}

\section{Additional Implementation Details}
\label{supp:implement_detail}
\subsection{Data Collection} 
As the collected web videos contain noisy content, we further explain the details of our data clearning strategies as follows. We filter out static videos based on extracted optical flow. The flow values are first normalized to the range of $[0,1]$, with an average flow value calculated for each frame. The videos that contain less than $10\%$ of frames with $> 0.1$ average flow value are considered static and removed from the dataset. 

\subsection{Inference Settings.}
We choose PerspectiveFields~\cite{perspectivefields} to estimate the elevation of per-frame and use LinearRegression to smooth the pitch sequence. Additionally, we modified the Venhancer~\cite{venhancer} to keep the $360^\circ$ close-loop property to super-resolution the output panorama video for a better 360 VR immersive experience in the webpage. Note that we do not use video SR of any kind in our comparison and ablation experiments.  

\subsection{Experiment Settings}
Due to space limit in the main paper, we further add explainations of our experiment settings. For human evaluation, we ask the users to evaluate the 360 videos across three dimensions:  graphics quality, structure plausibility, and temporal coherence. We provide both the 360 videos and its four perspective projection videos with $\phi=0, \theta=[0,90^\circ, 180^\circ,270^\circ]$. Graphics quality refers to the clarity and detail richness of the panorama and perspective video frames. Structure plausibility refers to the level of distortion in each perspective projections. Temporal coherence refers to the motion consistency and subject consistency: whether there's object suddenly appears or disappears, etc. 

For ablations, in the ablation on antipodal mask, we compare the model variant using both the antipodal mask and the directly-mapped mask with the model variant using only the directly-mapped mask. Note that we do not ablate on the directly mapped mask because its effectiveness in dual-design structure was already addressed in PanFusion~\cite{panfusion}. In the ablation on elevation-aware design ablation, we compare our full model using both the elevation sampling and elevation estimation with a model that do not use both of the designs to see the effect from elevation handling. 

% \subsection{Video Super-Resolution}
% 

% To split the supplementary pages from the main paper, you can use \href{https://support.apple.com/en-ca/guide/preview/prvw11793/mac#:~:text=Delete%20a%20page%20from%20a,or%20choose%20Edit%20%3E%20Delete).}{Preview (on macOS)}, \href{https://www.adobe.com/acrobat/how-to/delete-pages-from-pdf.html#:~:text=Choose%20%E2%80%9CTools%E2%80%9D%20%3E%20%E2%80%9COrganize,or%20pages%20from%20the%20file.}{Adobe Acrobat} (on all OSs), as well as \href{https://superuser.com/questions/517986/is-it-possible-to-delete-some-pages-of-a-pdf-document}{command line tools}.

\end{document}